%% file: biped.tex
\RequirePackage{amsmath}         %

\documentclass[letterpaper, 10pt, conference]{ieeeconf}      %

\IEEEoverridecommandlockouts                              %

\let\labelindent\relax                                    %

\usepackage{graphicx}            %
\usepackage{caption}
\usepackage{booktabs,caption}
\usepackage[flushleft]{threeparttable}
\usepackage{amssymb}             %
\usepackage{amsfonts}            %
\usepackage{enumitem}            %
\usepackage{multirow}            %
\usepackage{siunitx}             %
\usepackage{url}                 %
\usepackage{xspace}              %
\usepackage[T1]{fontenc}         %
\usepackage[hidelinks]{hyperref} %
\usepackage{wrapfig}
\usepackage{todonotes}
\usepackage{eurosym} 
\usepackage{balance}
\usepackage{cleveref}
\usepackage{hyperref}
\usepackage{tikz} %
\usepackage{subcaption} %
\usepackage{multirow}

\usepackage{pgfplots}
\pgfplotsset{compat=1.9}
\usepgfplotslibrary{groupplots}
\usetikzlibrary{matrix}

\graphicspath{{images/}}
\DeclareGraphicsExtensions{.pdf,.png,.jpg,.jpeg}

\newcommand{\biped}{AGILOped} %

\newcommand{\seclabel}[1]{\label{sec:#1}}

\newcommand{\figlabel}[1]{\label{fig:#1}}
\newcommand{\tablabel}[1]{\label{tab:#1}}
\newcommand{\eqnlabel}[1]{\label{eqn:#1}}

\newcommand{\figref}[1]{Fig.~\ref{fig:#1}\xspace}
\newcommand{\tabref}[1]{Table~\ref{tab:#1}\xspace}
\newcommand{\eqnref}[1]{(\ref{eqn:#1})\xspace}

\title{\LARGE \textbf{AGILOped: Agile Open-Source Humanoid Robot for Research}}

\author{Grzegorz Ficht, Luis Denninger, and Sven Behnke%
\thanks{All authors are with the Autonomous Intelligent Systems (AIS) Group, Computer Science Institute VI,
        University of Bonn, Germany. Email: {\tt\small ficht@ais.uni-bonn.de}.\newline
        This work has been supported by the German Federal Ministry of Education and Research (BMBF) grant 16ME0999 Robotics Institute Germany (RIG).}}
\usepackage{eso-pic}

\AtBeginDocument{\AddToShipoutPictureFG*{\AtTextUpperLeft{\put(0,\LenToUnit{9pt}){\parbox{\textwidth}{\centering\bfseries
10th IEEE International Conference on Advanced Robotics and Mechatronics (ARM), Portsmouth, UK, August 2025.
}}}}}
\begin{document}
\maketitle
\thispagestyle{empty}
\pagestyle{empty}

\begin{abstract}

With academic and commercial interest for humanoid robots peaking, multiple platforms are being developed.
Through a high level of customization, they showcase impressive performance. 
Most of these systems remain closed-source or have high acquisition and maintenance costs, however.
In this work, we present \biped~--- an open-source humanoid robot that closes the gap between high performance
and accessibility. Our robot is driven by off-the-shelf backdrivable actuators with high power density and uses standard electronic components.
With a height of 110\,cm and weighing only 14.5\,kg, \biped~can be operated 
without a gantry by a single person. Experiments in walking, jumping, impact 
mitigation and getting-up demonstrate its viability for use in research.

\end{abstract}

\input{1_introduction.tex}
\input{2_related_work.tex}
\input{3_hardware.tex} %
\input{4_electronicscontrol.tex} %
\input{5_evaluation.tex} %

\input{6_conclusions.tex}
\bibliographystyle{IEEEtran}
\balance
\bibliography{biped}

\end{document}

%% file: 1_introduction.tex
\section{Introduction}

Humanoid robotics is currently experiencing a surge in development, fueled by the elusive idea of universal automatons capable of performing work at a level similar to humans. This is the result of several key technologies maturing to the point of viability. Energy-dense and compact batteries and efficient motor controllers are able to quickly provide sufficient power to actuators, enabling powerful and dynamic motions.
The miniaturization of computing has enabled the embedding of online controls based on complex dynamics, as well as the use of Large Language Models. These have the ability to perform unstructured, unfamiliar tasks using multimodal input, evoking the sense of human reasoning.

Developing and integrating all these capabilities into a single platform is not trivial.
The needed multidisciplinary knowledge requires the collaboration of highly skilled professionals working in a dedicated, specialized environment. 
This comes at a cost, leaving development to a small number of players with the resources to sustain it. Such exclusivity of a universally applicable technology can have far-reaching socio-economic consequences, highlighting the need for accessible and open solutions. 

\begin{figure}[h!]
    \centering
    \begin{tikzpicture}
        \matrix (m) [matrix of nodes, nodes={inner sep=0.12cm, anchor=south west}, column sep=0, row sep=0] {
            \node {\includegraphics[height=4.8cm]{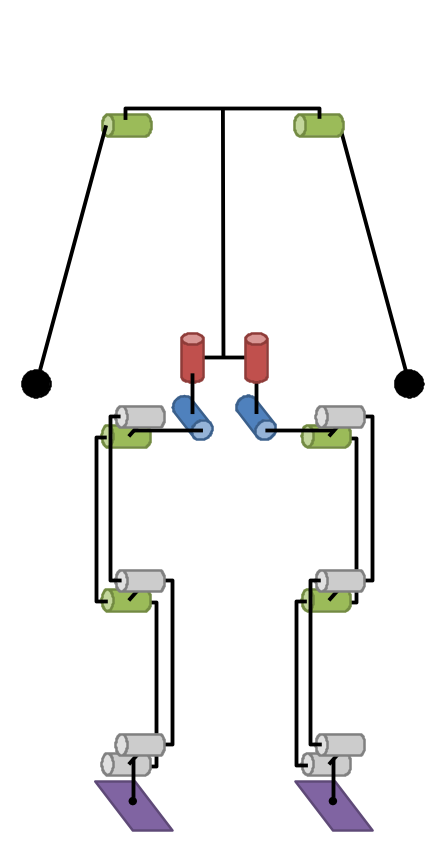}}; & 
            \node {\includegraphics[height=4.8cm]{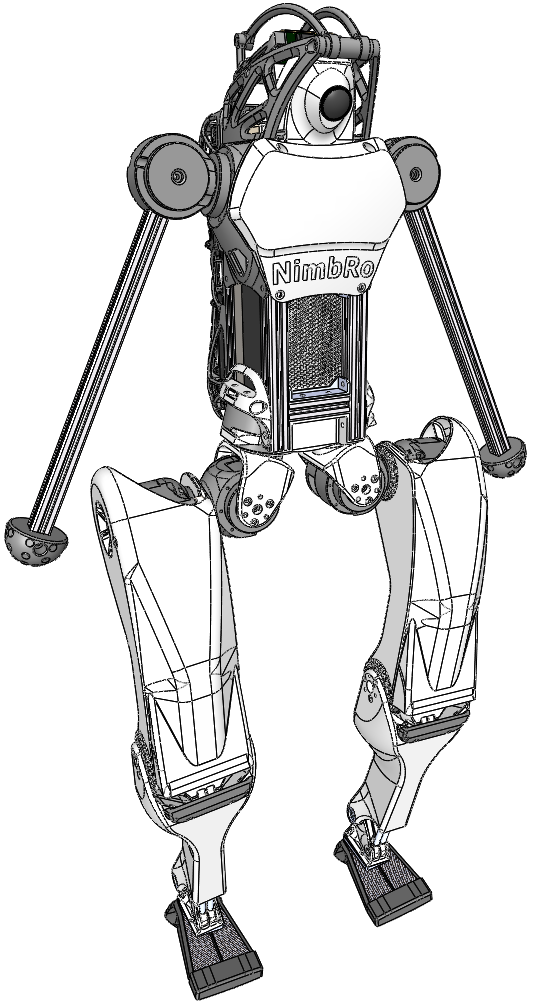}}; & 
            \node {\includegraphics[height=4.8cm]{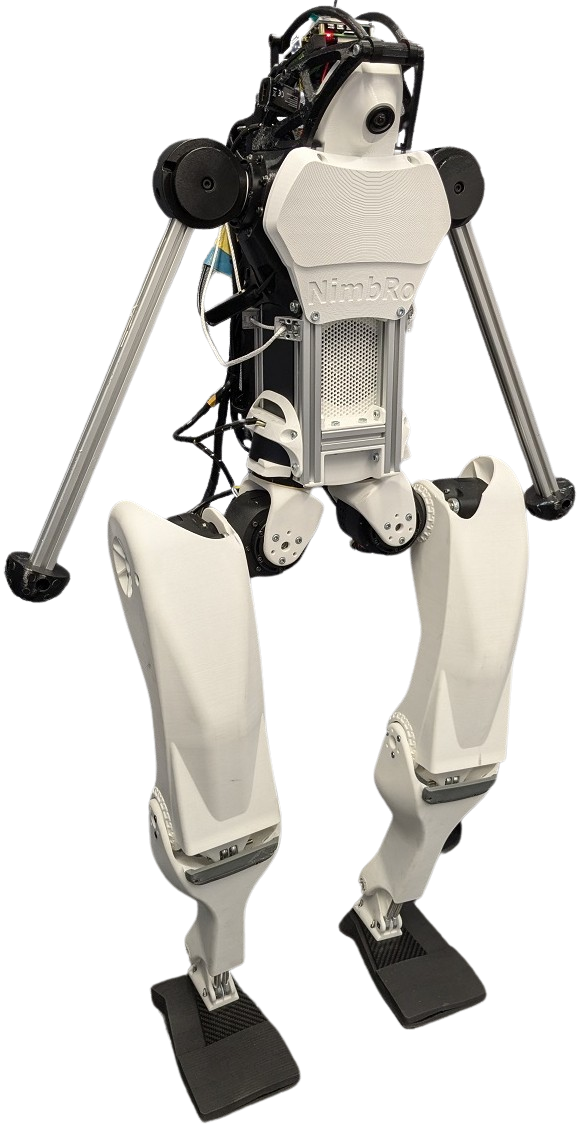}}; \\
        };
    \end{tikzpicture}\vspace{-0.2cm}
    \caption{The kinematics, CAD model and constructed version of \biped. With 10 active degrees of freedom (colored), a mostly 3D-printed structure, solely off-the-shelf components,
    and open-source design, our humanoid bridges the gap between accessible and agile platforms.}
    \figlabel{teaser}\vspace{-4ex}
\end{figure}

With \biped, shown in \figref{teaser}, we introduce a dynamically capable, open and affordable research platform. 
By relying only on a 3D printer, off-the-shelf components and basic tooling,
we greatly lower the entry barrier to humanoid robotics research. We describe the design 
principles, our hardware choices, and present results showcasing the robot's capabilities.

%% file: 2_related_work.tex
\section{Related Work}

Creating bipedal robots, able to reproduce the dynamic motions of humans has been a long-standing research 
interest. The MIT 3D Biped~\cite{playter1992control} was the first successful example capable of performing 
both untethered running and somersaults, a feat which hasn't been reproduced by a biped for the next 25 
years~\cite{atlasvid}. Running was also achieved by ATRIAS~\cite{atrias2016} with speeds exceeding 
\SI{9}{km/h}. Both robots are embodiments of the Spring-Loaded Inverted 
Pendulum (SLIP) model. Their hip-concentrated mass and low-inertia legs allow to achieve
explosive dynamic motions without need for complex controls, demonstrating the strength of 
a SLIP hardware template.

Performing such dynamic motions requires actuators not only capable of producing the necessary peak power,
but also capable of sustaining the repetitive high impact forces. While the hydraulics used in Atlas
offer extreme forces and naturally absorb impacts, they are the least user-friendly to work with due to 
noise and possible leakages. Furthermore, hydraulic actuators are not energy-efficient. Series-Elastic Actuators~(SEAs) are an alternative, as the embedded spring allows for energy storage and control through the relationship between force 
and displacement~\cite{negrello2015modular}. The spring limits the possible control bandwidth, though. Recent 
Quasi Direct Drive~(QDD) actuators with Brushless DC motors and integrated planetary gearbox do not suffer 
from this limitation. Originally developed for the Mini-Cheetah~\cite{katz2019mini}, they have been successfully 
applied to several recent humanoid designs~\cite{sim2022tello, li2023dynamic, artemis2023, saloutos2023design, liao2024berkeley}. 

\begin{table*}[t]
    \renewcommand{\arraystretch}{1.0}
    \caption{Comparison of established and recently developed humanoid platforms.}
    \tablabel{robot_comp}
    \centering
    \footnotesize
    \begin{threeparttable}
    \begin{tabular}{c | c c c c c c c c}
        \hline
        \textbf{Platform} & \textbf{Height} & \textbf{Weight} & \textbf{Price} & \textbf{Materials} & \textbf{Open Source} & \textbf{Off-the-shelf} & \textbf{Max. HFE}$^c$ & \textbf{Max. KFE}$^d$  \\
        & (cm) & (kg) & (USD) & (main) & (CAD) & & \textbf{Tor.} (Nm) & \textbf{Tor.} (Nm) \\
        \hline
        NAO~\cite{Gouaillier2009}                   & 57 & 4.5 & 14K & Plastic & Yes & No & 1.61$^{\text{b}}$ & 1.61$^{\text{b}}$ \\
        DARwIn-OP~\cite{Ha2011}                     & 45 & 2.8 & 12K & Plastic & Yes & Yes & 2.35$^{\text{b}}$ & 2.35$^{\text{b}}$ \\
        Booster T1~\cite{booster}\cite{robocuprobots}& 120 & 30 & 34K & Metal & No & No & - & 130 \\
        NimbRo-OP2x~\cite{ficht2020nimbro}          & 135 & 19 & 25K & Plastic & Yes & Yes & 40$^{\text{a,b}}$ & 80$^{\text{a,b}}$ \\
        Unitree H1~\cite{unitree2024g1}             & 130 & 35 & 90K & Metal & No & No & 88 & 139 \\
        MIT Humanoid~\cite{chignoli2021humanoid}    & 70 & 21 & - & Metal & No & No & 72 & 144 \\
        ARTEMIS~\cite{artemis2023}                  & 142 & 37 & - & Metal & No & No & 250 & 250 \\ 
        Unitree G1~\cite{unitree2024g1}\cite{robocuprobots}            & 130 & 35 & 16K$^{\text{e}}$-35K$^{\text{f}}$ & Metal & No & No & 88 & 139 \\
        Fourier GR-1~\cite{fourier2024gr1}\cite{robocuprobots}           & 165 & 60 & 200K & Metal & No & No & 230 & 230 \\
        Berkeley Humanoid~\cite{liao2024berkeley}   & 85 & 22 & 10K & Metal & No & No & 62.6 & 81.1 \\
        \hline
        \textbf{\biped~(ours)}                               & 110 & 14.5 & ~6.5K & Plastic & Yes & Yes & 40$^{\text{a}}$ & 80$^{\text{a}}$ \\
        \hline
    \end{tabular}
    \begin{tablenotes}
      \small
      \item \quad $^{\text{a}}${\scriptsize joint-configuration-dependent} \quad$^{\text{b}}${\scriptsize stall} \quad $^{\text{c}}${\scriptsize hip flexion/extension}
      \quad $^{\text{d}}${\scriptsize knee flexion/extension} \quad $^{\text{e}}${\scriptsize non-programmable} \quad $^{\text{f}}${\scriptsize for developers}
    \end{tablenotes}
    \end{threeparttable}\vspace{-3ex}
\end{table*}

Having resilient actuators might not always be sufficient to prevent hardware damage. 
As bipeds inevitably do fall, implementing various falling strategies is a viable option 
\cite{Qingqing2017fall, Ogata2007fall, goswami2014fall, kakiuchi2017}.
These strategies generally focus on damping the fall using the arms and bending the knees to reduce the 
fall height, as humans would. However, further design-related approaches, such as those explored by 
Wilken et al. \cite{wilken2009fall}, who designed compliant arms, have unfortunately been widely 
understudied. While recent robots have demonstrated improved dynamic motion capabilities, 
their ability to withstand repeated falls remains largely unexplored. A biomimetic approach
of enveloping a rigid structure with materials imitating soft tissues, might prove to be 
beneficial for reducing damage sustained from falling.

While it has been shown that humanoid platforms can achieve human-like dynamic motions, 
most remain inaccessible to many research groups. \tabref{robot_comp} provides an overview 
of recently developed platforms. The platforms cited above, coming from research groups,
such as the MIT Humanoid~\cite{saloutos2023design}, Berkeley Humanoid~\cite{liao2024berkeley}, 
HECTOR~\cite{li2023dynamic} and ARTEMIS~\cite{artemis2023}, are unavailable to those interested.
Because they use custom components, their results are not reproducible by other researchers.
Although commercially available humanoids are on the horizon, e.g. Unitree G1~\cite{unitree2024g1}, 
Fourier-GR1~\cite{fourier2024gr1}, and Booster T1~\cite{booster}, they come with their drawbacks. 
They are either expensive, make the user dependent on the manufacturer, or come with limited 
source-code access. This typically restricts full control, 
particularly at the low level, making these platforms less attractive for research purposes. 
Alternatively, the NimbRo-OP2X~\cite{ficht2020nimbro} platform or iterations of the DARwIn-OP~\cite{Ha2011}
offer a more research-friendly solution. These systems provide access to freely available 
CAD models and utilize off-the-shelf actuators, allowing for full control over hardware 
and simplifying maintenance. However, their limited motor specifications or small size prevent them from achieving 
the dynamic motions of modern humanoids, highlighting the need for a dynamic and accessible platform.

%% file: 3_hardware.tex
\section{Design}
\seclabel{hardware_design}

The design of our robot is intended to provide a low entry point into humanoid robotics development without 
compromising on the ability to perform dynamic tasks. To achieve this, the platform must be universally accessible, 
which includes obtaining the robot at an affordable price, effortless and reliable operation, easy maintenance, 
and customization possibilities. An overview of the hardware specifications of \biped\,is given in \tabref{biped_specs}. 
With a height of \SI{110}{cm} and weight of \SI{14.5}{kg}, it matches the 
measurements of a typical 6-year-old child. The size was intentionally chosen for 
its balance of meaningful interaction and safe operation within human-scale environments.
We leverage our experience with 3D-printed humanoids~\cite{ficht2017nop2,ficht2020nimbro} to make \biped\,
accessible and easily manufactured using only a commercially available 3D printer and basic tools. This enables
modularity, as parts can quickly be altered to have improved physical properties, house newer 
components or implement novel functionality. The design heavily relies on selective compliance, efficiently 
combining the rigidity of aluminum and flexibility of 3D-printed plastics and polyurethanes of varying hardness 
to deliver cohesive, minimalistic and robust robot hardware. 

For the design of \biped, we utilized off-the-shelf components to significantly simplify manufacturing and 
maintenance. Off-the-shelf items are cost-effectively produced at scale, extensively tested by the manufacturer, 
well-documented and ready to use. With only 10 actuators controlling 12 joints, we emphasize simplicity
and cost-effectiveness assuring that the robot is approachable to users with varying levels of experience. 
Through careful selection of modular components and materials embedded in a light-weight design, we have achieved 
a price point of 6,380 USD, making \biped\ the most affordable and lightest humanoid in its size class. To promote 
collaboration and reduce exclusivity within the field of humanoid robotics, the design files were open-sourced online\footnote{\url{https://nimbro.net/Humanoid/AGILOped}}.

\begin{table}[h]
\renewcommand{\arraystretch}{1.2}
\setlength{\tabcolsep}{2pt}
\caption{\biped\, specifications.}
\tablabel{biped_specs}
\centering
\footnotesize
\begin{tabular}{c c c}
\hline
\textbf{Type} & \textbf{Specification} & \textbf{Value}\\
\hline
\multirow{4}{*}{\textbf{General}} & Height \& Weight & \SI{110}{cm}, \SI{14.5}{kg}\\
& Battery & 2$\times$LiPo (\SI{26.1}{V}, \SI{4.5}{Ah})\\
& Battery life & \SI{1.5}{}--\SI{2.5}{h}\\
& Material & Aluminum, PLA/Nylon, TPU\\
\hline
\multirow{5}{*}{\textbf{Computing}}
& Base Controller & RaspBerry Pi 3B+ \\
& Specs & 4$\times$\SI{1.4}{GHz}, \SI{1}{GB} RAM\\
& \multirow{2}{*}{I/O} &\vspace{-2pt}  4$\times$USB 2.0, GPIOs, 1$\times$HDMI, \\
& &  Ethernet, Wi-Fi \\
& (Optional) & NVIDIA Jetson \\
\hline
\multirow{6}{*}{\textbf{Actuators}} 
& Total & 10$\,\times\,$MyActuator RMD X6-40\\
& Torque & \SI{18}{Nm}~(Rated), \SI{40}{Nm}~(Peak) \\
& No load speed (48V) & \SI{11.5}{rad/s} \\
& Feedback & Position, Velocity, Torque\\
& Encoder (Motor) & 16\,bit/rev\\
& Communication & CAN (1Mbit) \\
\hline

\multirow{8}{*}{\textbf{Cost~(USD)}} 
& Actuators & 10$\,\times\,$545 \\
& Batteries & 2$\,\times\,$150 \\
& IMU & 1$\,\times\,$30 \\
& Raspberry Pi 3B+ & 1$\,\times\,$35 \\
& PiCAN2Duo & 1$\,\times\,$70 \\
& Misc Hardware & 1$\,\times\,$495 \\
& NVidia Jetson & 1$\,\times\,$(200 -- 2000) \\
& \textbf{Total} & 6,380~(6,580 -- 8,380) \\
\hline
\end{tabular}\vspace{-3ex}
\end{table}

\subsection{Actuator Module and Power}
Originally custom-developed as a novel actuator for the Mini-Cheetah~\cite{katz2019mini}, Quasi Direct Drives~(QDD) 
with brushless DC motors and integrated planetary transmission offer high torque density and control at high bandwidths.
Through in-house development one might have more control over aspects of the design process, but it is associated 
with significant costs in time and effort through development, testing and manufacturing. Derivatives of these 
actuators have since become widely used in legged robotics and available from numerous vendors 
in various sizes~\cite{tmotor,westwood,myactuator}. We opt to use the MyActuator RMD X6-40 with \textit{two-stage} 
planetary gearing for all joints, mainly for its high torque within a compact, modular form-factor~(see \tabref{biped_specs}). 

Because of the direct correlation of motor diameter and torque, it may seem counter-intuitive to opt for a 
low-diameter motor. The X6-40 compensates for this by a 1:36 planetary gearbox, provide up to \SI{40}{Nm} torque. 
This is not detrimental to proprioception, due to the smaller radius keeping the 
rotor inertia low~(\SI{28.8}{kg\,cm^2}). It is lower than that of a large-diameter, flat 
X10-40 actuator~(\SI{39.7}{kg\,cm^2}), with single-stage 1:7 planetary gearbox and similar 
peak torques of \SI{40}{Nm}, but twice the weight~(\SI{1.1}{kg})~\cite{myactuator}.

Despite the extra reduction stage, X6-40 can reach angular velocities exceeding \SI{10}{rad/s},
which is equivalent to peak joint velocities of a human sprinting faster than \SI{6}{m/s}~\cite{belli2002moment}.
For the same output speed, the motor turns faster---compared to actuators with lower gear 
ratio---which improves motor efficiency and, thus, reduces thermal load.
Although actuators in a similar weight class are capable of providing higher no-load speeds~(e.g. Unitree A1), 
their cost is roughly double that of the X6-40, while providing only \SI{80}{\%} of its peak torque.
Each X6-40 actuator has an integrated motor controller, equipped with a high-resolution 16-bit encoder mounted 
on the motor shaft. The chosen actuators operate with voltages in the range 20--52\,V in position, velocity, torque and 
impedance control mode. Commands and feedback are transmitted over a CAN bus. 

To operate the actuators at the maximum possible velocity, we use two \SI{26.1}{V} UAV batteries connected 
in series for a total supply voltage of \SI{52.2}{V}. The high-power density, protection features, rugged casing, 
smart switch and charge manager remove the necessity to design specialized electronics around raw cells. 
We take advantage of this by implementing a hot-swap circuit 
based on two high-power diodes (as seen in \figref{physical_layer}). In normal operation, the diodes are reverse-biased and
the batteries are directly connected to the motors. When any of the batteries is removed for a swap, the 
diode closes the circuit. The simplicity and cost-effectiveness of this setup comes with the requirement
that all components need to tolerate a wide voltage input.
\subsection{Leg Design}

Each leg features a hybrid, serial-parallel kinematic chain with five joints and biologically inspired elements.
The leg joints are powered through four collocated actuators: three at the hip~(yaw, roll, pitch) and one at the knee~(pitch). 
All axes in the hip intersect at a common point to simplify the kinematics. The outward hip pitch placement aims to
mimic the bicondylar angle of the femoral shaft naturally acquired by humans, which is known to increase their efficiency. This is due to 
a part of the load being distributed to the skeletal structure, but also by allowing the feet to be placed closer 
to the midline---minimizing lateral movement~\cite{bramble2004endurance,tayton2007femoral,tardieu2010development}. 
In the sagittal plane, we employ double 4-bar parallelograms~(actuated at the $q_{th}$ thigh and $q_{sh}$ shank) 
that control the hip $q_{h}$, knee $q_{k}$ and ankle $q_{a}$: 
\begin{equation} \eqnlabel{jointrel}
q_h = q_{th},\quad q_k = q_{sh} - q_{th},\quad q_a = -q_{sh}
\end{equation}
First, this reduces the number of actuators~(thus the inertia) and achieves a completely passive ankle joint $q_a$ 
with the foot orientation horizontally constrained to the torso. The second benefit is a faster knee, and torque $\tau_k$
shared between thigh and shank. This is defined by the mapping Jacobian \emph{\textbf{J}}, obtained through differentiating~\eqnref{jointrel}:
\begin{equation}\eqnlabel{jacobians}
\begin{bmatrix} \dot{q}_h\\ \dot{q}_k\\ \dot{q}_a \end{bmatrix} = \emph{\textbf{J}}\begin{bmatrix}\dot{q}_{th}\\ \dot{q}_{sh}\end{bmatrix},\; 
\begin{bmatrix} \tau_h\\ \tau_k\\ \tau_a \end{bmatrix} = \emph{\textbf{J}}\begin{bmatrix}\tau_{th}\\ \tau_{sh}\end{bmatrix},\;
\emph{\textbf{J}}=\begin{bmatrix}1 &0\\-1 &1\\0 &-1 \end{bmatrix}.
\end{equation}
The parallelograms are neatly hidden within the 3D-printed structure, and despite their presence we are able to 
achieve a high range of motion~(see \figref{legrom}).

\begin{figure}[t]
    \centering
    \centering
    \includegraphics[width=0.84\linewidth]{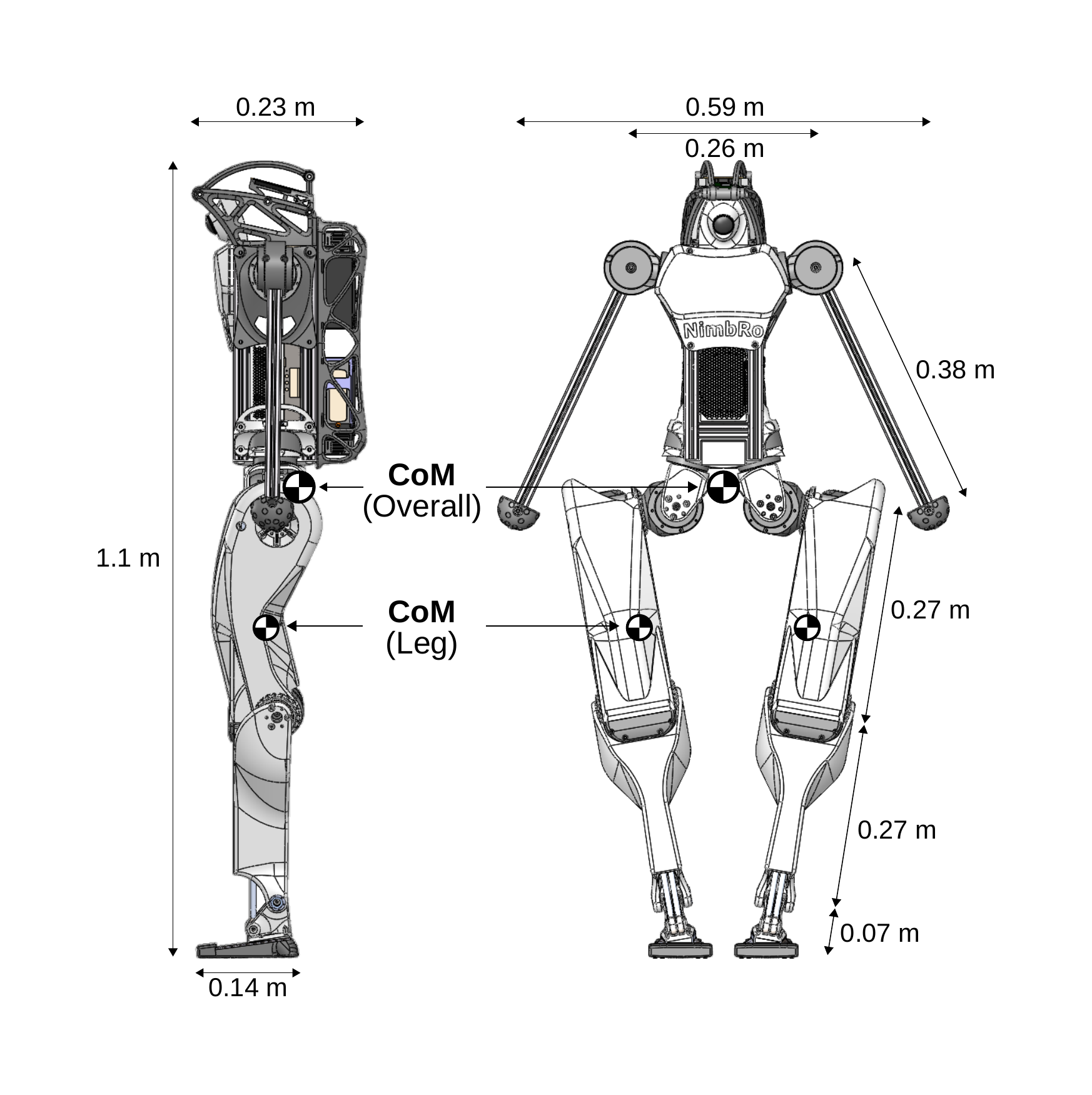}\vspace{-0.5cm}
    \caption{\biped~dimensions labelled in frontal and side view.}
    \figlabel{robotdimensions}
\end{figure}

\begin{figure}[b]\figlabel{legrom}
    \centering
    \centering
    \begin{subfigure}[b]{0.48\linewidth}
        \centering
        \includegraphics[height=6.0cm]{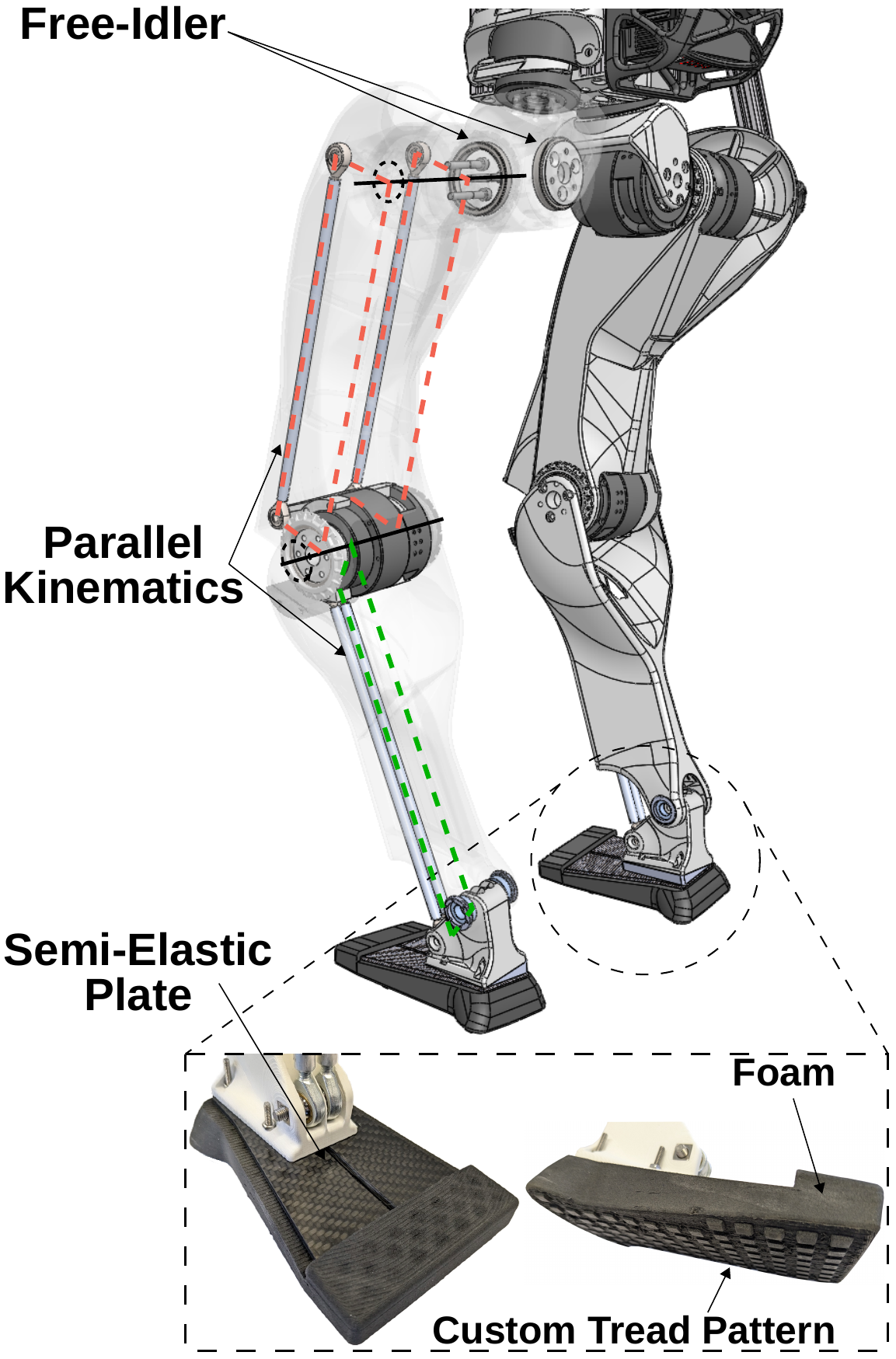}
        \caption{Parallel kinematics and foot design.}
        \label{fig:pk_vis}
    \end{subfigure}
    \hfill
    \begin{subfigure}[b]{0.48\linewidth}
        \centering
        \includegraphics[height=6.0cm]{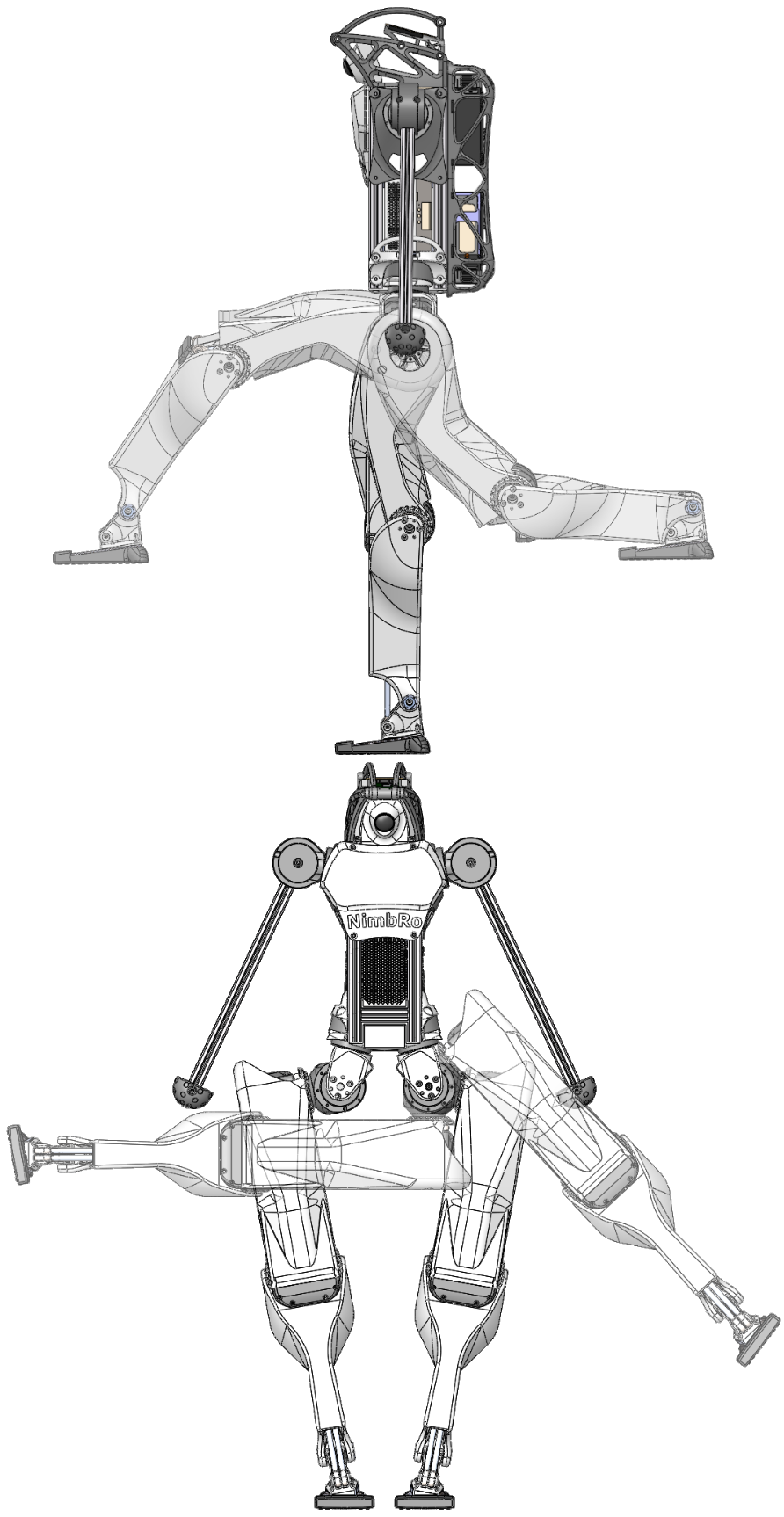}
        \caption{Range of motion}
        \label{fig:rom_vis}
    \end{subfigure}
    \caption{Leg design details.}
    \figlabel{legrom}
\end{figure}

As almost the complete leg is 3D-printed, achieving rigidity without significant weight poses 
a challenge. Although CNC-milled structural components are not usually burdened with these issues, 
they come at a higher cost and require a production process that is more demanding in expertise, effort and costs.
We reinforce the structure through modulating thickness, ribbing, and strategic use of fasteners.
\biped\,was prototyped and operated successfully using parts made with Polylactide (PLA), however we recommend 
using Nylon instead due to its durability and heat resistive properties. This approach is
experimentally proven to not hinder the long-term structural integrity, as our Nylon-printed NimbRo-OP2(X)~\cite{ficht2017nop2}\cite{ficht2020nimbro}
robots have operated for several years in tough RoboCup conditions without breaking any of the structural parts since their construction~\cite{pavlichenko2023robocup}. 
As the actuators lack a free-idler on the opposite end of the actuation axis, we implemented compact idlers
mimicking the driving side of the actuator. The connection between the hip roll and pitch actuator mounts uses
three screws, which simultaneously reinforce the structure and lock the hip idler bearing in place.
We close the parallel kinematic chains with manually-threaded aluminum rods, with screwed-in universal 
joints. They are doubled and symmetrically spaced for improved axis definition, uniform loading, and redundancy,
greatly enhancing the resistance to buckling and deformations.

Bipeds with 4-DoF legs typically have point-feet, which makes them statically unstable. They need to rely on 
dynamic stability to maintain balance~\cite{ghansah2024dynamic}. Our implementation not only decreases inertia, 
but allows us to use an elongated foot to achieve static stability while standing. The parallel kinematic ankle pitch joint poses
a challenge, as it requires the foot to have a certain level of adaptability to the ground while simultaneously
being sufficiently rigid to support the robot. The implemented design draws inspiration from biology~\cite{frund2022guideline}
and the world of sports~\cite{ortega2021energetics}. Elastic plates are angled and directly connected 
to the ankle to provide the foot with longitudinal stiffness and energy storage capabilities. By separating 
the plates, we mimic the medial and lateral arches, achieving lateral compliance. The plates are inserted
into a sole, which is 3D-printed using a unique Thermoplastic Polyurethane (varioShore TPU)~\cite{iacob2023effect} with a foaming agent.
The material behaves like an elastomer below a threshold value of the printing temperature. Passing it however,
makes it foam and lowers the shore hardness. The final foot design is sufficiently rigid, compliant and 
comes with an embedded tread pattern for increased contact friction. 

\begin{figure}[b]
    \centering
    \includegraphics[width=0.86\linewidth]{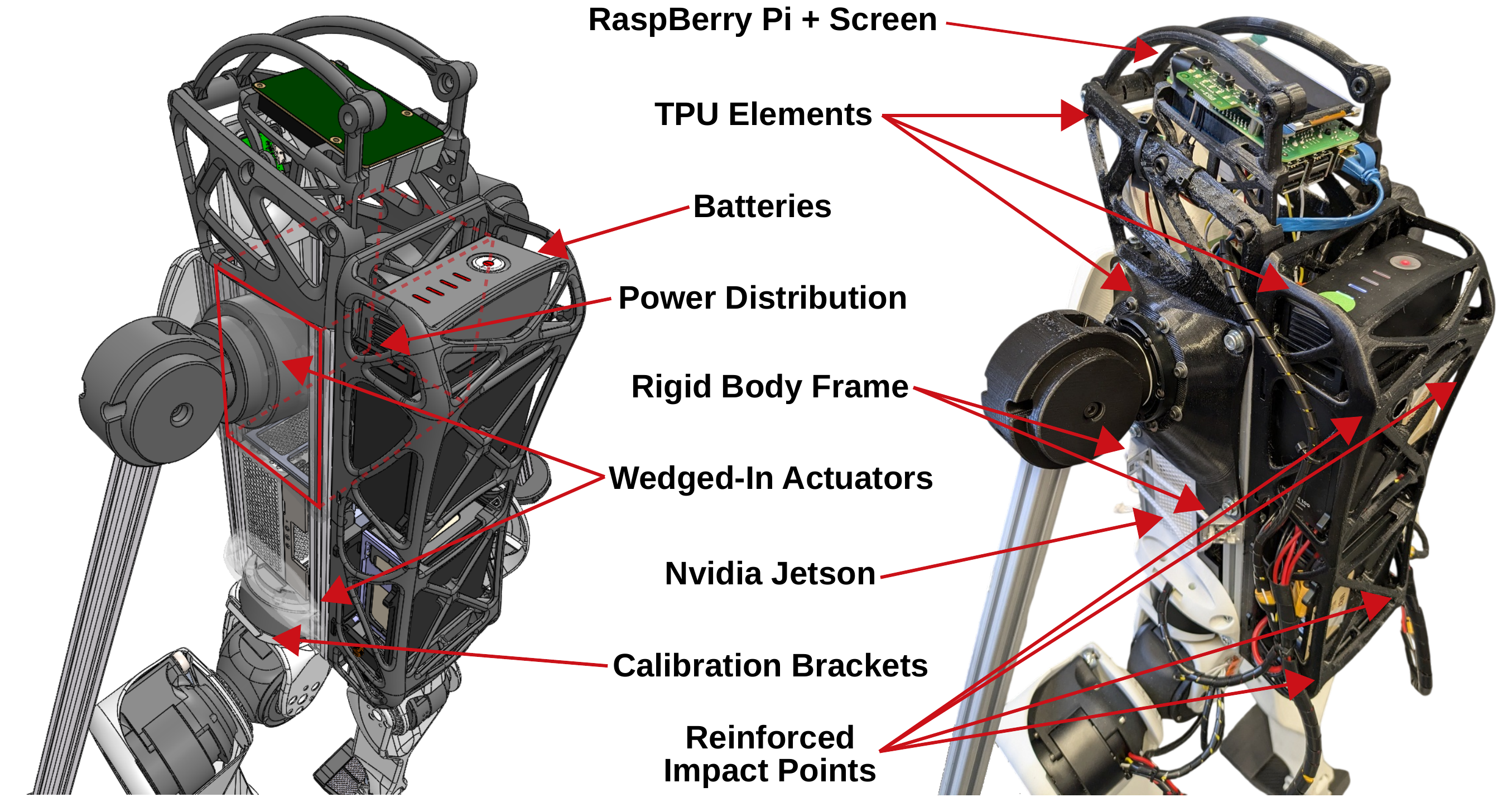}
    \caption{Upper body design details. Featuring a rigid cage, elastomer-based impact mitigation, embedded computers, batteries, and power distribution.}
    \figlabel{cage}
\end{figure}

In total, the movable part of the leg (excluding the hip yaw within the torso) weighs \SI{3.51}{kg}, with 
actuators totaling \SI{1.77}{kg}. The collocated structure greatly simplifies the mechanical design 
without negatively impacting the inertia, as all actuators but the knee actuator are tightly packed within the hip. 
Furthermore, platforms which do possess ankle actuators~\cite{liao2024berkeley}\cite{artemis2023}, 
place them below the knee with a similar contribution to the reflected inertia. 
The lower part of the leg, comprised of the shank, ankle, foot and 
parallel rods sums up to \SI{0.56}{kg}, leaving \SI{84}{\%} of the weight next to the hip. In the most extended 
configuration, the leg center of mass is only \SI{0.16}{m} away from the origin of the hip, which translates to 
\SI{25.8}{\%} of the leg length~(See \figref{robotdimensions}). Such characteristics with low rotational inertia 
should favor high-bandwidth control, as well as being beneficial for controllers relying on reduced-order models.

\subsection{Upper Body Design}

The upper body is comprised of the torso and two 1-DoF arms. The frame of the torso is built from 
aluminum extrusions arranged to form a cage, providing the majority of the rigidity. The actuators snugly 
fit between the extrusions, and lock into place through 3D-printed brackets. Given the rigidity of the aluminum 
frame, a plastic bracket is sufficient to support the mounting of the hips and maintain leg rigidity. Additionally,
the hips have an alignment bracket, which allow the legs to collapse into a defined position for calibration. 
Robots prioritizing locomotion research do not necessarily need to incorporate arms.
The basic arms are expected to aid \biped\, in surviving falls and performing get-up motions,
which provides a level of functionality above the platforms that do not have them altogether~\cite{liao2024berkeley}\cite{xia2024duke}.
Furthermore, with \biped's open-source nature we aim for users to refine and adapt the arm design for specific tasks.
To increase impact resistance without deteriorating the performance, we employ selective compliance around 
the rigid frame. For this reason, the shoulder mounts are printed using TPU. This provides the shoulder with sufficient 
lateral compliance to mitigate falls while allowing rigid actuation in the pitch axis. 
The torso cage also features a \textit{backpack} comprised of two symmetric battery compartments.
Each of these is made from an off-the-shelf aluminum housing, press fit into a 3D-printed TPU enclosure.
Strategic enforcing contact points and placement of supporting struts enhanced the elasticity 
in dispersing impact energy, already proving itself useful when working on \biped's development.

The frame also has ample space to house the computing, sensing units, and power distribution.
A Raspberry Pi with a screen is used for basic system control and mounted at the top of the 
robot for visual feedback. This unit, along with an optional camera form a simple robot \textit{head}. 
They are protected with a 3D-printed TPU cage. Inside the torso, there is space for the power 
distribution and an optional NVidia Jetson computing unit~(see \figref{cage}) . 
With this, we aim for \biped\, to support modern (learning-based and model-based) control approaches 
and cater to a wide range of users.

%% file: 4_electronicscontrol.tex
\section{Control Architecture}
\seclabel{electronics}
\subsection{Physical Layer}
\begin{figure}[t!]\figlabel{electronics}
    \centering
    \includegraphics[angle=-90,origin=c,width=0.83\linewidth]{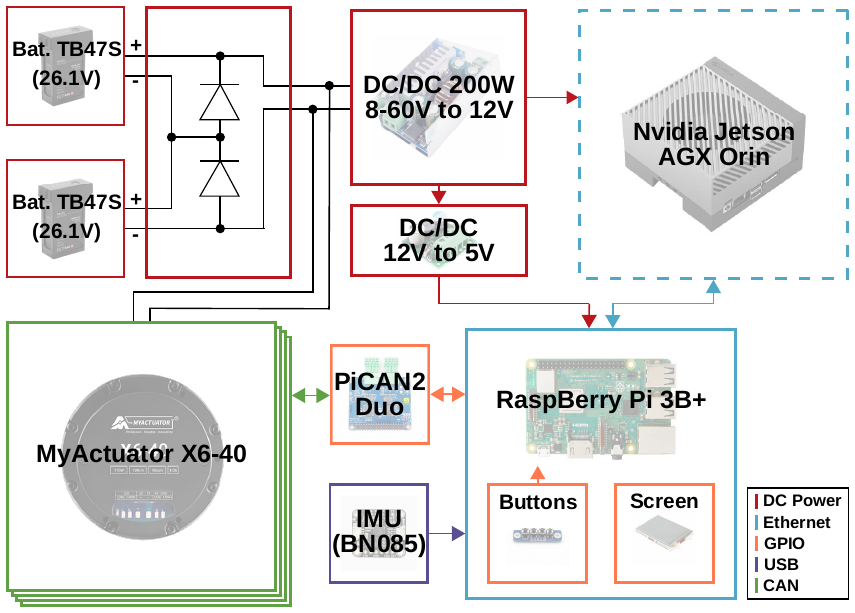}
    \vspace{-1cm}
    \caption{Electronics layout.
            \biped~is powered by two \SI{26}{V} batteries, providing a combined peak voltage of 52V to the motors. 
            A wide-input DC/DC converter steps down the voltage to \SI{12}{V} for the Jetson AGX Orin.
            Voltage is further converted into \SI{5}{V} for the Raspberry Pi, serving as the base controller and CAN bus master. 
            For higher-requirement applications, the Jetson AGX Orin can connect to the Raspberry Pi via Ethernet.}
    \figlabel{physical_layer}
\end{figure}

At the core of our control scheme lies a Raspberry Pi 3B+ Single Board Computer (SBC), 
which we chose due to its versatility, community-driven software, and long-term manufacturing support. 
It's form factor and compatibility allow for effortless upgrades to e.g. the more recent Raspberry Pi 5. 
A multitude of add-on boards~(\textit{hats}) exist that greatly enhance its capabilities. \biped's setup 
(see ~\figref{physical_layer}) extends the SBC with a screen for user feedback, buttons for basic controls and a PiCAN2 
Duo hat to enable actuator communication over two separate CAN buses.

An Inertial Measurement Unit~(IMU) is essential in giving the robot a sense of spatial motion.
By fusing the 3D measurements of an accelerometer, gyroscope and magnetometer, an Attitude and 
Heading Reference System~(AHRS) allows for precise orientation feedback. Due to the required knowledge 
in sensor calibration, implementation and tuning of fusion algorithms, obtaining a low-latency and precise 
estimate proves to be a challenge. Current dynamically capable systems have to rely on solutions in the 
\SI{1000}{USD} price range~\cite{katz2019mini}~\cite{artemis2023}. For \biped\, we have developed a 
plug-and-play, low-cost AHRS solution, based on RP2040 microcontroller and BNO085 AHRS modules. 
The BNO085 has built-in sensor calibration and fusion algorithms and various modes of output, 
providing the user directly with quality orientation feedback with rates up to \SI{1}{kHz}. 
To further ease the usage, we implemented USB communication through the \texttt{rosserial} library, which
de/serializes ROS messages and provides straightforward ROS integration. Our open-source implementation 
has a total cost of \SI{30}{USD} and is available online \footnote{\url{https://github.com/gficht/imu_rosserial/}}.
In combination with the kinematic model and joint measurements, we obtain accurate estimates 
of the system dynamics~\cite{ficht2023centroidal}.

\subsection{Low-Level Control}
The Raspberry Pi operates using Ubuntu Mate OS, with ROS support and serves as the low-level controller, 
which we refer to as \textit{RosPi} controller. Two CAN buses manage communication with five motors each, 
with a baud-rate limiting the communication to \SI{4}{kHz}, resulting in a maximum rate of \SI{800}{Hz}
per motor. Due to the high frequency of communications and low-latency (measured at~\SI{1.7}{ms}), we treat
the motors as torque sources. 

To enable compatibility with our position-control-based framework, impedance
controllers are implemented. The target torque $\tau$ is computed from position $q$ and velocity 
$\dot{q}$ errors and their associated stiffness $K_P$ and damping $K_D$ gains,
added to a feedforward term $\tau_{ff}$ and limited to a maximum allowable value $\tau_{max}$:
\begin{equation}\label{eq:imp_control}
    \tau = min(K_P ({q}_{set}-q) + {K}_D ({\dot{q}_{set}}-\dot{q})+\tau_{ff}, {\tau_{max}}).
\end{equation}
As the motors have absolute position encoders only on the motor shaft, the output position
of the joint needs to be inferred. This is done by placing the robot into a pre-defined calibration pose,
achieved by making the robot collapse his legs and resting the thighs on the hip brackets (see \figref{cage}).

Several software safety mechanisms have been implemented to protect both the robot and its surroundings. 
If a motor loses connection for more than $\SI{100}{ms}$, its internal controller will stop it.
Additionally, if communications to the higher-layer node is lost, a shutdown command is sent to the motors.
We have also implemented continuous monitoring of motor temperatures and voltages, their values 
on the Raspberry Pi screen. Persistent high temperatures, over-voltage, or torques 
exceeding typical values trigger warnings and shutdown the motors to prevent damage.

\begin{figure}[tbh]
\parbox{\linewidth}{\centering
\includegraphics[width=1.0\linewidth]{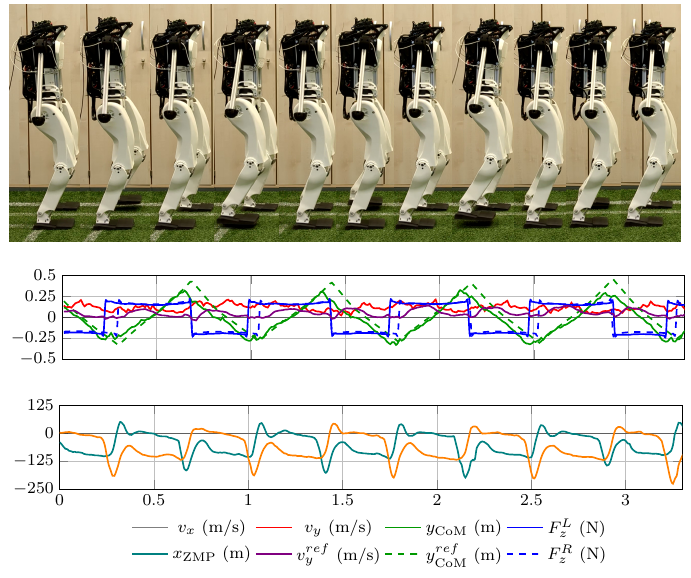}}\vspace*{-1ex}
    \caption{Forward balanced walking with an omnidirectional, feedback-enhanced, 
    CPG-based gait. (top) motion sequence. (bottom) CoM and ZMP series, and noticeable contact force exchanges 
    between the feet.}
    \label{walkingplot}
\end{figure}

\subsection{Higher-Level Control}
\biped's higher-level controls are built on our open-source, ROS-based software stack that leverages the modularity 
of ROS. First introduced in 2013, the framework has undergone continuous improvements while maintaining 
its core structure. The main strength being modularity, allowing the integration of custom hardware 
architectures, gaits, and motion controllers through plug-ins. Motion sequences can be easily designed 
using keyframes through the built-in trajectory editor. This design also enables seamless integration 
with simulators like Gazebo and MuJoCo. The software stack has proven its versatility and adaptability 
across various platforms, including the Igus Humanoid Open Platform~\cite{allgeuer2016igus}, the 
NimbRo-OP2X~\cite{ficht2020nimbro}, and now \biped. 

%% file: 5_evaluation.tex
\section{Evaluation} 
\subsection{Bipedal Walking}

Bipedal locomotion is one of the most fundamental skills for humanoid robots. For the experiments, we initially 
adapted and tuned the feedback-enhanced Central Pattern Generated~(CPG) gait used on the NimbRo-OP2X~\cite{pavlichenko2023robocup}.
The robot is tasked with walking forward on a compliant artificial grass surface. In the lateral plane, 
the robot compares its current and reference Linear 
Inverted Pendulum~(LIP) Center of Mass~(CoM) position and velocity states~\cite{kajita1991study}. By using 
closed-form predictions of the end-of-step state, the gait frequency is adjusted to either delay or accelerate 
a step. Sagittally, a regulator that computes a desired CoM velocity from ZMP and CoM tracking errors, steers 
the CoM towards stability. The tracking performance, along with a time series of the gait can be observed 
in Fig.~\ref{walkingplot}. One noteworthy feature is the inclusion of contact forces obtained directly from 
joint torques and limb Jacobian, showcasing the proprioceptive capabilities of the actuators.
\subsection{Jumping}

\begin{figure}[tbh]
    \parbox{\linewidth}{\centering
\includegraphics[width=1.0\linewidth]{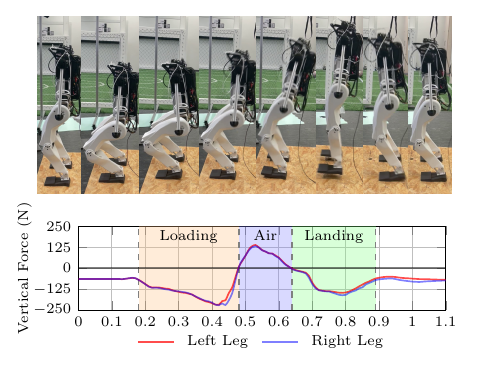}}\vspace*{-1ex}
    \caption{Vertical jump experiment. (top) Sequence of keyframes that propel the center of 
    mass upwards and perform the landing. (bottom) Contact forces estimated using torque measurements and Jacobian.}
    \label{jumpingplot}
\end{figure}

\begin{figure*}[t]
   \centering
   \centering
   \begin{tikzpicture}
       \matrix (m) [matrix of nodes, nodes={inner sep=-0.01cm, anchor=south west}, column sep=0, row sep=0] {
           \node {\includegraphics[height=3cm]{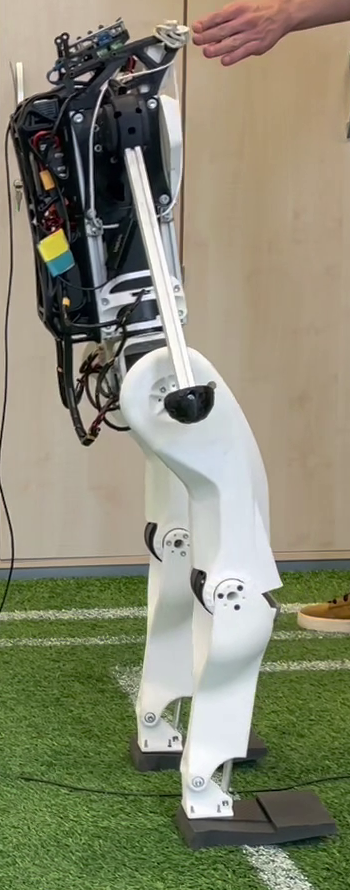}}; & 
           \node {\includegraphics[height=3cm]{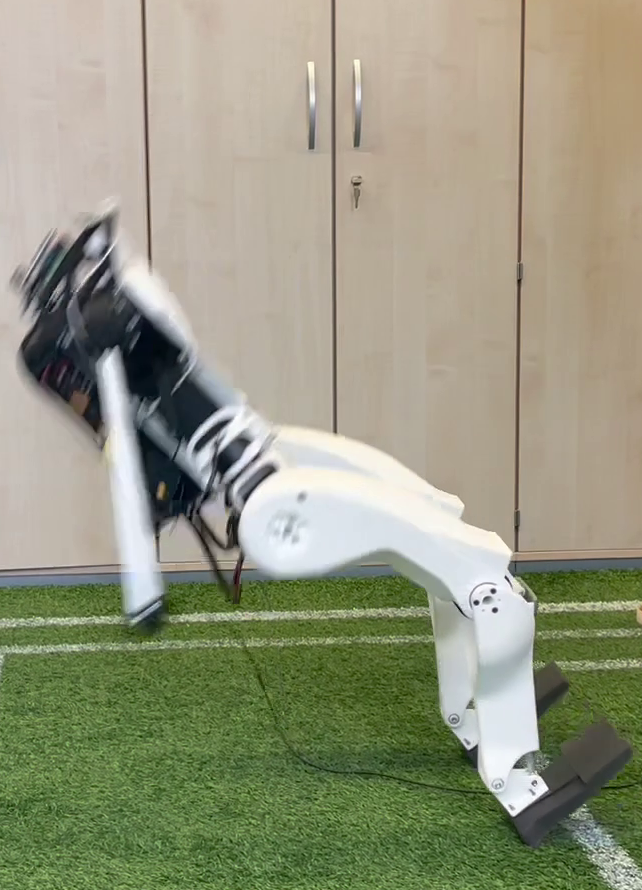}}; & 
           \node {\includegraphics[height=3cm]{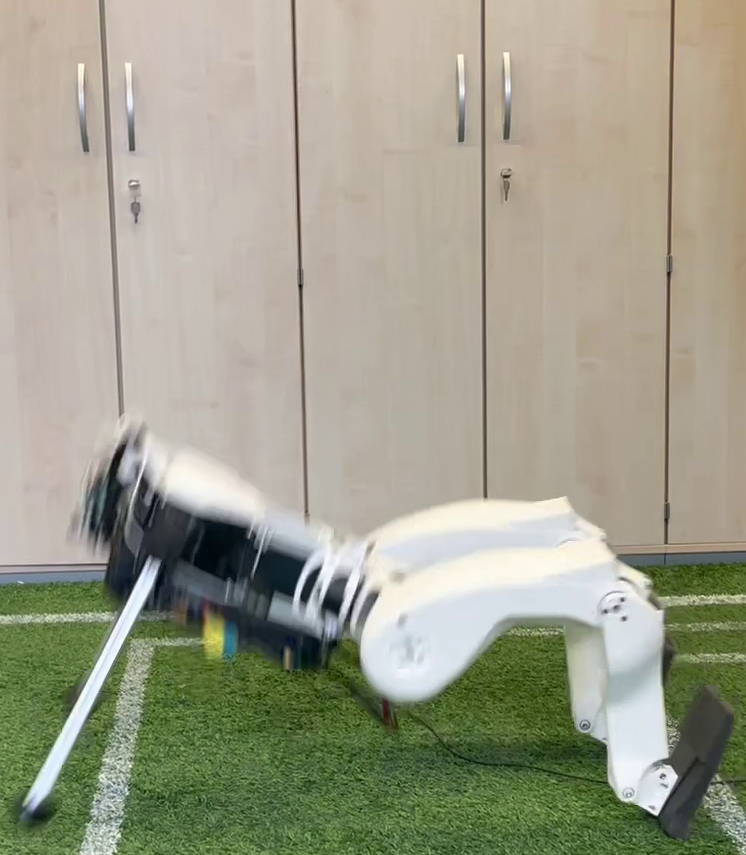}}; & 
           \node {\includegraphics[height=3cm]{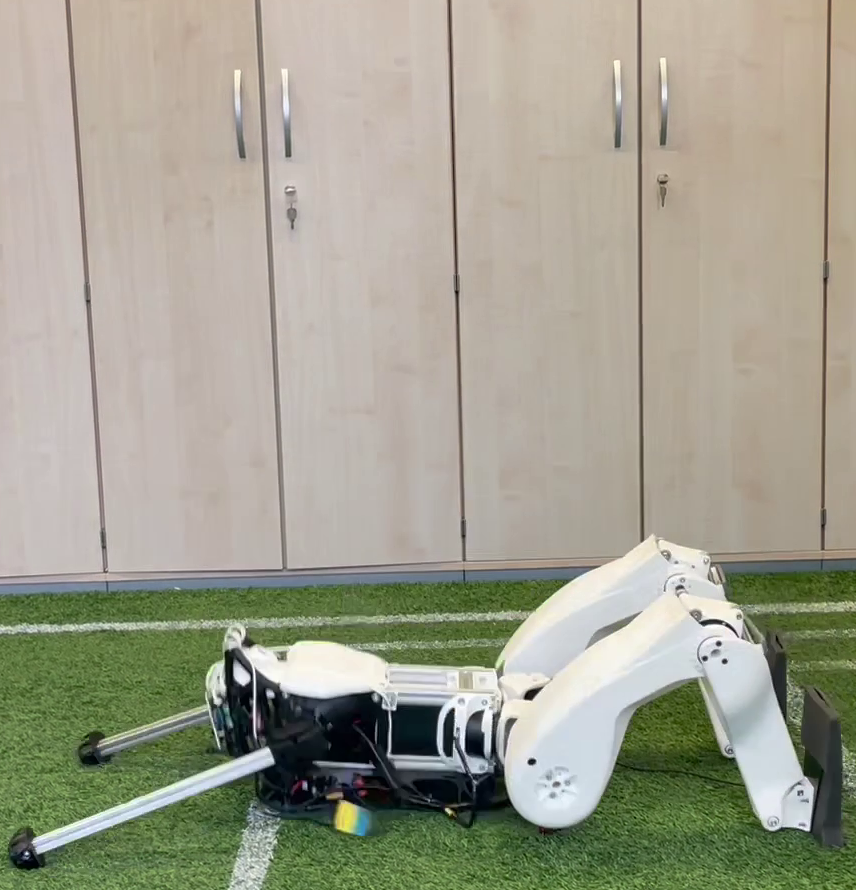}}; & 
           \node {\includegraphics[height=3cm]{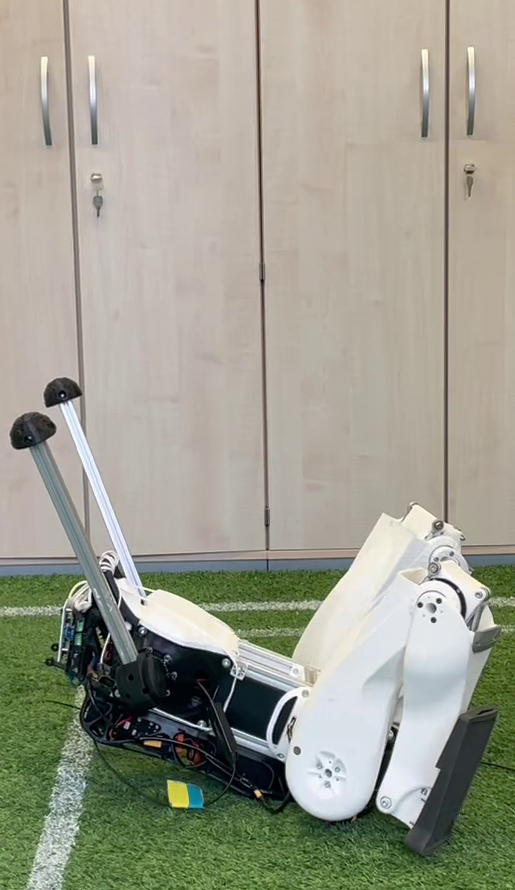}}; & 
           \node {\includegraphics[height=3cm]{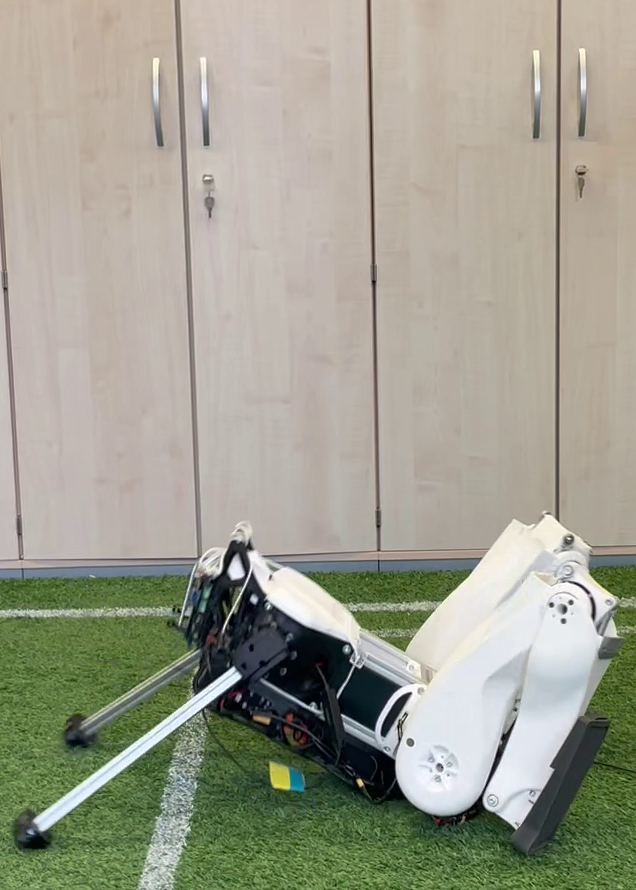}}; & 
           \node {\includegraphics[height=3cm]{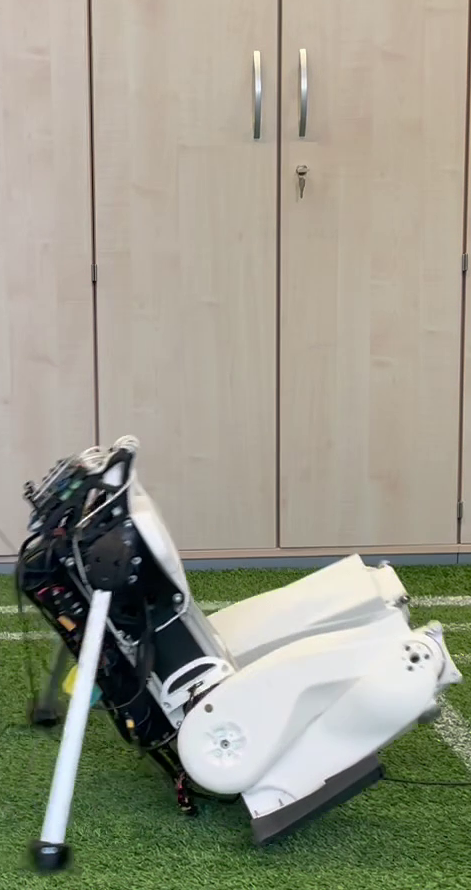}}; & 
           \node {\includegraphics[height=3cm]{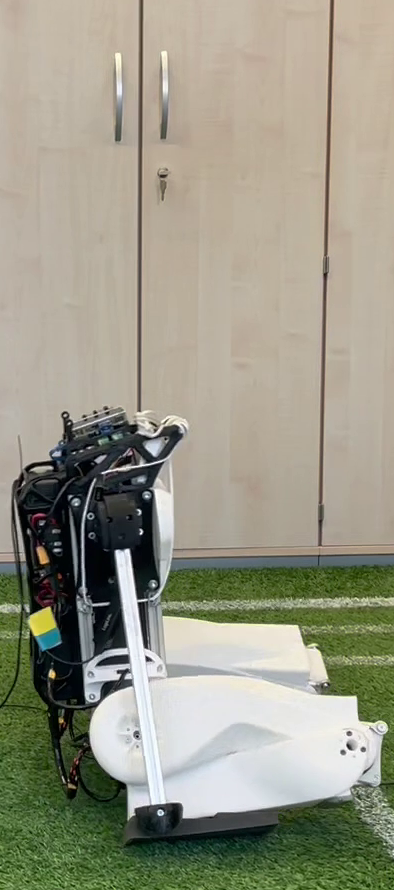}}; & 
           \node {\includegraphics[height=3cm]{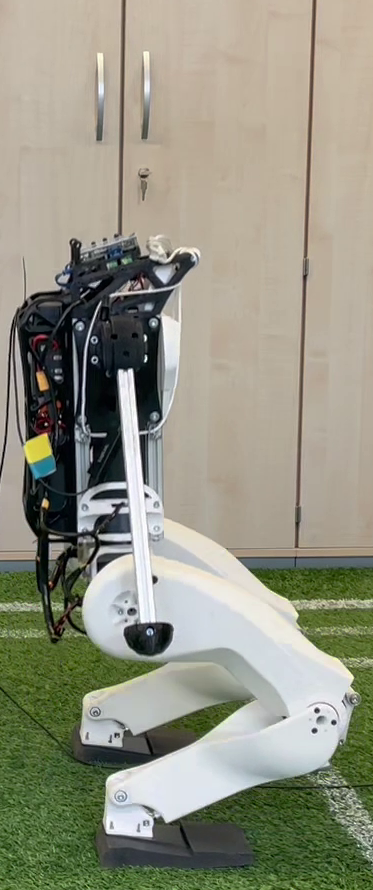}}; & 
           \node {\includegraphics[height=3cm]{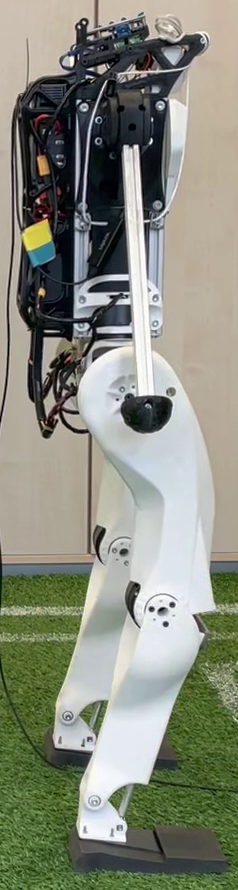}}; \\
       };
   \end{tikzpicture}
   \caption{Time series of \biped\, falling backwards and standing up again. Upon detecting the fall, the robot rapidly moves its arms backward to reduce the impact.
   From a prone position, it pushes itself back up onto its feet, followed by a standing-up motion.}
    \figlabel{falling_plot}\vspace{-3ex}
\end{figure*}

By simply designing a set of whole-body position keyframes, we are able to swiftly move the center of mass
and propel the robot upwards, as shown in Fig.~\ref{jumpingplot}. The motions are not optimal, and merely serve as a means 
to display the peak power capabilities of the actuators. Without a dedicated force controller, we are unable to 
meaningfully apply full torque. Hence, we can reliably achieve jumps of only about \SI{10}{cm}. Most likely due to incorrect
CoM assumptions, \biped\,jumps slightly backwards, wasting some of the propelling force. 
\subsection{Falling Mitigation and Standing-up}
To test the robustness of the hardware with respect to impacts, we intentionally push the robot from any side 
causing it to fall, as illustrated in \figref{falling_plot}.
When a fall is detected, \biped\, responds by swiftly moving its arms 
and bracing for impact. The stiffness of the impedance controllers is greatly reduced, while damping is kept at a noticeable
level. This significantly breaks the fall, allowing the robot to gently bounce off on the elastic battery backpack.

Following the fall, \biped\, immediately initiates the get-up routine by retracting its legs and performing a strong push-off.
This propels it back onto the feet from which getting up is straightforward, demonstrating the combined effect
of the achieved actuator and structural compliance.

%% file: 6_conclusions.tex
\section{Conclusions}

In this work, we have presented the hardware and software design of our humanoid robot \biped. 
The combination of off-the-shelf backdrivable proprioceptive actuators with high power density, standard electronics, and effective material usage within 
a light-weight and minimalistic design has led to the creation of a capable and affordable research platform. 
The mechanical and electrical simplicity make \biped~accessible for novices and 
experts alike. 

We reported experiments on walking, jumping, falling mitigation, and getting-up.
We hope that other researchers will notice the potential of \biped. Its open hard- and software will enable them to contribute their
developments. We envision \biped~to perform robust agile locomotion together with dynamic whole-body motions like kicking.